\documentclass[sigconf]{acmart}

\AtBeginDocument{%
  \providecommand\BibTeX{{%
    \normalfont B\kern-0.5em{\scshape i\kern-0.25em b}\kern-0.8em\TeX}}}

\setcopyright{none}

\acmConference[Preprint]{Preprint}{arXiv}{Code: https://github.com/CyberAgentAILab/cmaes}

\usepackage{latexsym}
\usepackage{amsmath} 

\usepackage{mathtools}       
\usepackage{stmaryrd}        
\usepackage{bm}              

\providecommand{\ind}[1]{\mathbb{I}\{#1\}} 
\renewcommand{\geq}{\geqslant} 
\renewcommand{\leq}{\leqslant} 
\DeclarePairedDelimiterX{\inner}[2]{\langle}{\rangle}{#1, #2}

\usepackage{amsthm}
\usepackage{xcolor}

\usepackage{ifthen}
\usepackage{cleveref}

\theoremstyle{definition}

\usepackage{algorithm}
\usepackage{algpseudocode}

\algnewcommand\algorithmicset{\textbf{Set:}}
\algnewcommand\Set{\item[\algorithmicset]}

\makeatletter
\let\OldStatex\Statex
\renewcommand{\Statex}[1][3]{%
  \setlength\@tempdima{\algorithmicindent}%
  \OldStatex\hskip\dimexpr#1\@tempdima\relax}
\makeatother

\newcommand{\textbfit}[1]{\textbf{\textit{#1}}}
\newcommand{\code}[1]{\colorbox[gray]{0.95}{\texttt{#1}}}

\usepackage{listings}

\definecolor{codegreen}{rgb}{0,0.6,0}
\definecolor{codegray}{rgb}{0.5,0.5,0.5}
\definecolor{backcolour}{rgb}{0.99,0.99,0.99}

\definecolor{brightmaroon}{rgb}{0.76, 0.13, 0.28}
\definecolor{dukeblue}{rgb}{0.0, 0.0, 0.61}
\definecolor{pakistangreen}{rgb}{0.0, 0.4, 0.0}

\lstdefinestyle{mystyle}{
    backgroundcolor=\color{backcolour},   
    commentstyle=\color{pakistangreen},
    keywordstyle=\color{brightmaroon},
    numberstyle=\tiny\color{codegray},
    stringstyle=\color{dukeblue},
    basicstyle=\ttfamily\footnotesize,
    breakatwhitespace=false,         
    breaklines=true,                 
    captionpos=b,                    
    keepspaces=true,                 
    numbers=left,                    
    numbersep=5pt,                  
    showspaces=false,                
    showstringspaces=false,
    showtabs=false,                  
    tabsize=2,
    frame=single
}
\usepackage{subfig}

\newcommand{\Qmark}{\spacefactor=1000\relax?}
\begin{document}
\title{\texttt{cmaes}: A Simple yet Practical Python Library for CMA-ES}

\author{Masahiro Nomura}
\email{nomura@comp.isct.ac.jp}
\orcid{0000-0002-4945-5984}
\authornote{Masahiro Nomura and Masashi Shibata contributed equally to this work.
This work was conducted while they were affiliated with CyberAgent.}
\affiliation{%
  \institution{Institute of Science Tokyo}
  \city{Yokohama}
  \state{Kanagawa}
  \country{Japan}
  \postcode{226-0026}
}

\author{Masashi Shibata}
\email{mshibata@preferred.jp}
\orcid{0009-0000-8656-5122}
\authornotemark[1]
\affiliation{%
  \institution{Preferred Networks}
  \city{Chiyoda}
  \state{Tokyo}
  \country{Japan}
  \postcode{100-0004}
}

\author{Ryoki Hamano}
\email{hamano_ryoki_xa@cyberagent.co.jp}
\orcid{0000-0002-4425-1683}
\affiliation{%
  \institution{CyberAgent}
  \city{Shibuya}
  \state{Tokyo}
  \country{Japan}
  \postcode{150-0042}
}

\begin{abstract}
The covariance matrix adaptation evolution strategy (CMA-ES) has been highly effective in black-box continuous optimization, as demonstrated by its success in both benchmark problems and various real-world applications.
To address the need for an accessible and powerful tool in this domain, we developed \texttt{cmaes}, a simple and practical Python library for CMA-ES.
\texttt{cmaes} is characterized by its simplicity, offering intuitive use and high code readability.
This makes it suitable for quick use of CMA-ES, as well as for educational purposes and seamless integration into other libraries.
Despite its simple design, \texttt{cmaes} maintains advanced functionality.
It incorporates recent advancements in CMA-ES, such as learning rate adaptation for challenging scenarios, transfer learning, mixed-variable optimization, and multi-objective optimization capabilities.
These advanced features are accessible through a user-friendly API, ensuring that \texttt{cmaes} can be easily adopted in practical applications.
We present \texttt{cmaes} as a strong candidate for a practical Python CMA-ES library aimed at practitioners.
The software is available under the MIT license at 
\textcolor{blue}{\url{https://github.com/CyberAgentAILab/cmaes}}.
\end{abstract}



\maketitle

\section{Introduction}
\label{sec:intro}


Black-box optimization focuses on optimizing an objective function whose internal structure is either unknown or inaccessible.
In this approach, the objective function is regarded as a ‘‘black-box’’:
while it provides outputs in response to given inputs, the internal process generating these outputs remains concealed during optimization.
This approach is invaluable when the objective function is complex, costly to evaluate, or lacks an explicit analytical form.
The primary challenge in black-box optimization is enhancing the quality of the objective function value with minimal evaluations.

The covariance matrix adaptation evolution strategy (CMA-ES)~\cite{hansen2001completely,hansen2016cma} stands out as a method for solving black-box continuous optimization problems.
It optimizes by sampling candidate solutions from a multivariate Gaussian distribution, enabling parallel solution evaluations and showcasing significant parallelism.
Its superiority is evident when compared to other black-box optimization methods, particularly in challenging scenarios such as ill-conditioned, non-separable, or rugged problems~\cite{rios2013derivative}.
The effectiveness of CMA-ES spans various real-world applications, including computer vision~\cite{kikuchi2021constrained,kikuchi2021modeling}, natural language processing~\cite{sun2022black,sun2022bbtv2}, reinforcement learning~\cite{ha2018world}, model merging~\cite{akiba2025evolutionary}, test-time adaptation~\cite{niu2024test}, and automated machine learning~\cite{nomura2021warm,purucker2023cma}.

In this work, we introduce \texttt{cmaes}, a Python-implemented CMA-ES library rapidly gaining popularity in the field.
This is evidenced by over 450 GitHub stars and its integration into renowned Python libraries such as Optuna~\cite{akiba2019optuna} and Katib~\cite{george2020scalable}.

The cornerstone of \texttt{cmaes} lies in its \emph{simplicity} and \emph{practicality}:
To ensure \emph{simplicity}, our focus has been on high code readability, making it an inspiration for other evolution strategy libraries such as evojax~\cite{tang2022evojax} and evosax~\cite{lange2023evosax}, a useful educational resource.
Regarding its \emph{practicality}, \texttt{cmaes} stands out by incorporating recent significant advancements in CMA-ES, setting it apart from other libraries.
This integration includes (i)~automatic learning rate adaptation for solving \emph{difficult} problems such as multimodal and noisy environments \emph{without} expensive hyperparameter tuning, (ii)~transfer learning, (iii)~mixed-variable optimization, and (iv)~multi-objective optimization.
These advanced features are easily accessible through user-friendly APIs, establishing \texttt{cmaes} as the primary choice for practitioners.

\begin{figure*}[htbp]
\centering
\includegraphics[width=0.97\textwidth,trim=5 5 5 5,clip]{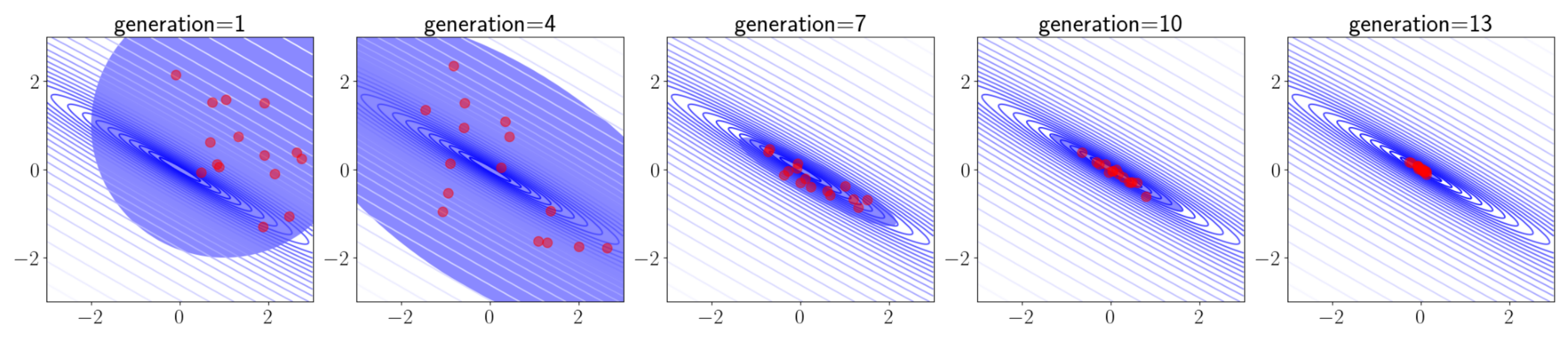}
\caption{Example of CMA-ES optimizing $f(x) = f_{\rm Ellipsoid}(R x)$, where $f_{\rm Ellipsoid}(x) = x_1^2 + (10 x_2)^2$ and $R \in \mathbb{R}^{2 \times 2}$ is a rotation matrix rotating $\pi / 6$ around the origin.
Population size $\lambda = 15$ and the initial distribution $m^{(0)} = [1.0, 1.0], \sigma^{(0)} = 1.0, C^{(0)} = I$. 
Ellipse represents the distribution of CMA-ES and the red points represent the sampled solutions.
CMA-ES efficiently addresses ill-conditioned and non-separable problems by adapting the distribution parameters.
}
\label{fig:cmaes-illustration}
\end{figure*}

In this paper, we aim to provide a comprehensive overview of \texttt{cmaes}'s motivation and distinctive features, making this paper a valuable guide for practitioners to effectively use the library.
The remainder of this section discusses related work on Python software for CMA-ES.
Section~\ref{sec:bg} explains the specifics of CMA-ES adopted in this study.
Section~\ref{sec:cma_oss} discusses the design philosophy behind \texttt{cmaes}, provides basic usage information, and discusses the software aspects of our library.
Section~\ref{sec:recent} discusses several recent advances in the field of CMA-ES that can be used through \texttt{cmaes}.
Section~\ref{sec:conclusion} concludes the paper with a summary and discussion.

\paragraph{\textbf{Related Work}.}
While there are several sophisticated libraries for CMA-ES, pycma~\cite{pycma} is particularly renowned for its comprehensive features and detailed documentation.
Indeed, pycma includes several features that our library lacks, such as handling of nonlinear constraints~\cite{atamna2017linearly}, sophisticated restriction of the covariance matrix~\cite{akimoto2014comparison,akimoto2016projection,akimoto2020diagonal}, and surrogate-assisted methods~\cite{hansen2019global}.
However, it might pose challenges for users seeking an in-depth understanding due to its complexity.
In contrast, our library \texttt{cmaes} focuses on basic and essential features, with an emphasis on simplicity and ease of understanding, appealing to practitioners who prefer straightforward implementations.
This simplicity also enhances the library's flexibility for integration of \texttt{cmaes} with other libraries.

Beyond pycma, the Python ecosystem houses other notable CMA-ES implementations.
For example, evojax~\cite{tang2022evojax} and evosax~\cite{lange2023evosax} are JAX-based libraries, ideal for leveraging GPUs or TPUs in scalable optimization.
pymoo~\cite{blank2020pymoo} specializes in multi-objective optimization, which is only partially covered by our library.
Additionally, Nevergrad~\cite{nevergrad} offers a diverse range of optimization methods, facilitating comparative studies between methods such as CMA-ES and Bayesian optimization~\cite{frazier2018tutorial}.
This variety underscores the rich landscape of optimization tools available, each with its unique advantages and specialized applications.

\section{CMA-ES}
\label{sec:bg}

We consider minimization\footnote{This is w.l.o.g. because replacing $f$ with $-f$ allows to consider maximization.} of the objective function $f : \mathbb{R}^d \to \mathbb{R}$.
CMA-ES optimizes $f$ by sampling solutions from a multivariate Gaussian distribution $\mathcal{N}(m, \sigma^2 C)$, wherein $m \in \mathbb{R}^d$ is the mean vector, $\sigma \in \mathbb{R}_{>0}$ is the step-size, and $C \in \mathbb{R}^{d \times d}$ is the covariance matrix.
Following the evaluation of solutions for $f$, CMA-ES updates the distribution parameters $m$, $\sigma$, and $C$ to produce more promising solutions.
Figure~\ref{fig:cmaes-illustration} provides an example of optimization using CMA-ES.
In this paper, we present a version of Ref.~\cite{hansen2016cma} that our library \texttt{cmaes} adopts.
For a detailed explanation of CMA-ES, see the Hansen's excellent tutorial~\cite{hansen2016cma}.

We briefly summarize CMA-ES to clarify the exact variant implemented in \texttt{cmaes}.
CMA-ES begins by initializing the parameters $m^{(0)}, \sigma^{(0)}$, and $C^{(0)}$ parameters.
The following steps are then repeated until the stopping condition is met:

\noindent
{\bf Step 1. Sampling and Ranking}\\
In the $g + 1^{\rm st}$ generation (with $g$ starting from $0$), for the population size $\lambda$, 
$\lambda$ candidate solutions $x_i\ (i=1, 2, \cdots, \lambda)$ are independently sampled from $\mathcal{N}(m^{(g)}, ( \sigma^{(g)} )^2 C^{(g)})$:
\begin{align}
    y_i &= \sqrt{ C^{(g)} } z_i, \\
    x_i &= m^{(g)} + \sigma^{(g)} y_i,
\end{align}
where $z_i \sim \mathcal{N}(0, I)$ and $I$ is the identity matrix.
These solutions are evaluated using the function $f$ and sorted in ascending order.
We denote $x_{i:\lambda}$ as the $i$-th best candidate solution $f(x_{1:\lambda}) \leq f(x_{2:\lambda}) \leq \cdots \leq f(x_{\lambda:\lambda})$.
Additionally, $y_{i:\lambda}$ represents the random vectors corresponding to $x_{i:\lambda}$.

\noindent
{\bf Step 2. Update Evolution Path}\\
Using the parent number $\mu \leq \lambda$ and the weights $w_i$, where $w_1 \geq w_2 \cdots \geq w_{\mu} > 0$ and $\sum_{i=1}^{\mu} w_i = 1$, the weighted average $dy = \sum_{i=1}^{\mu} w_i y_{i:\lambda}$ is calculated.
The evolution paths are then updated by:
\begin{align}
    p_{\sigma}^{(g+1)} &= (1-c_{\sigma}) p_{\sigma}^{(g)} + \sqrt{c_{\sigma} (2-c_{\sigma}) \mu_w} \sqrt{ C^{(g)} }^{-1} dy, \\
    p_{c}^{(g+1)} &= (1-c_{c}) p_{c}^{(g)} + h_{\sigma}^{(g+1)} \sqrt{c_{c} (2-c_{c}) \mu_w} dy,
\end{align}
where $\mu_w = 1 / \sum_{i=1}^{\mu} w_i^2$, $c_{\sigma}$, and $c_c$ are the cumulation factors, and $h_{\sigma}^{(g+1)}$ is the Heaviside function,
\begin{align}
    h_{\sigma}^{(g+1)} = \begin{cases}
1 & {\rm if}\ \frac{\| p_{\sigma}^{(g+1)} \|}{\sqrt{1 - (1-c_{\sigma})^{2(g+1)}} \mathbb{E}[\| \mathcal{N}(0, I) \|]} < 1.4 + \frac{2}{d+1}, \\
0 & {\rm otherwise}.
\end{cases}
\end{align}
where $\mathbb{E}[\| \mathcal{N}(0, I) \|] \approx \sqrt{d} \left( 1 - \frac{1}{4d} + \frac{1}{21 d^2} \right)$ is the expected norm of a sample of a standard Gaussian distribution.

\noindent
{\bf Step 3. Update Distribution Parameters}\\
The distribution parameters are updated as follows:
\begin{align}
    &m^{(g+1)} = m^{(g)} + c_m \sigma^{(g)} dy, \\
    &\sigma^{(g+1)} = \sigma^{(g)} \exp \left( 1, \frac{c_{\sigma}}{d_{\sigma}} \left( \frac{\| p_{\sigma}^{(g+1)} \|}{\mathbb{E}[\| \mathcal{N}(0, I) \|]} - 1 \right) \right) , \label{eq:stepsize} \\
    &C^{(g+1)} = \left( 1 + (1-h_{\sigma}^{(g+1)}) c_1 c_c (2-c_c) \right) C^{(g)} \nonumber \\
    &\ + c_1 \underbrace{\left[ p_{c}^{(g+1)} \left( p_{c}^{(g+1)} \right)^{\top} - C^{(g)} \right]}_{\text{rank-one update}} + c_{\mu} \underbrace{\sum_{i=1}^{\lambda} w_i^{\circ} \left[ y_{i:\lambda} y_{i:\lambda}^{\top} - C^{(g)} \right]}_{\text{rank-$\mu$ update}},
\end{align}
where $w_i^{\circ} := w_i \cdot (1 \text{ if } w_i \ge 0 \text{ else } d / \| \sqrt{C^{(g)}}^{-1} y_{i:\lambda} \|^{2} )$,
$c_m$ is the learning rate for $m$;
$c_1$ and $c_{\mu}$ are the learning rates for the rank-one and -$\mu$ updates of $C$, respectively; and
$d_{\sigma}$ is the damping factor for the $\sigma$ adaptation.
The update of distribution parameters in CMA-ES is closely related to the natural gradient descent~\cite{akimoto2010bidirectional}.

CMA-ES is invariant to order-preserving transformations of the objective function, utilizing solution rankings instead of actual values.
Additionally, it possesses affine invariance within the search space.
These features enable the generalization of its empirical successes to a wider range of problems~\cite{hansen2013principled}.

\section{\texttt{cmaes}: Simplicity $\&$ Practicality}
\label{sec:cma_oss}

This section aims to offer a comprehensive explanation of the various software aspects of our library.
Initially, we delve into the core design philosophy of our library in Section~\ref{sec:design_philosophy}, where we outline the fundamental principles that have shaped its development.
Following this, Section~\ref{sec:getting_started} provides guidance on the installation process and the basic usage of the library, ensuring a smooth start for new users.
We then shift our focus to features specifically designed to enhance the software's quality: these include the methodology for identifying unexpected errors via fuzzing in Section~\ref{sec:fuzzing}, and the animated visualization tools in Section~\ref{sec:vis}.
To demonstrate the practical application of \texttt{cmaes} in other libraries, Section~\ref{sec:optuna} presents a use case involving Optuna~\cite{akiba2019optuna}, showcasing the integration and utility of \texttt{cmaes} in a real-world context.

\subsection{Design Philosophy}
\label{sec:design_philosophy}
Our ultimate goal is to develop a Python CMA-ES library that serves as a strong practical choice for real-world users.
To this end, the library is designed with simplicity and practicality as its guiding principles.

\paragraph{\textbf{Simplicity}.}
To achieve our goal, the library is designed with utmost simplicity. We identify three major benefits of this approach:
Firstly, simplicity enhances software quality, easing debugging and maintenance, which is crucial for long-term reliability and practical use. Simplified maintenance also invites more feedback from the developer community, leading to a more refined product.
Secondly, a more lightweight library is inherently more user-friendly. A straightforward API makes it easier for users to understand and utilize the library effectively.
Lastly, from an educational standpoint, a simpler implementation aids in understanding the code in relation to the algorithm. This makes the library accessible not only to knowledgeable practitioners but also to beginners eager to learn about CMA-ES.

\paragraph{\textbf{Practicality}.}
While we prioritize keeping the library simple, it is essential to develop a library that solves the real-world challenges practitioners face.
To maximize the library's applicability in real-world settings, it is important to integrate insights from cutting-edge research on CMA-ES, particularly those with a strong practical motivation.
However, it is equally important to ensure the library is user-friendly, as not all practitioners are knowledgeable about CMA-ES.
The challenge we address is ensuring the practicality while maintaining the simplicity of our library.

\paragraph{\textbf{Our Library}.}
Based on the above principles, we have designed the library as follows:

To ensure \textbfit{simplicity}, our primary focus has been on achieving high code readability in the implementation.
Although the code is not included in this paper due to space limitations, renowned evolution strategy libraries such as evojax~\cite{tang2022evojax} and evosax~\cite{lange2023evosax} are inspired by our implementation.\footnote{\url{https://github.com/google/evojax/blob/67c90e1/evojax/algo/cma_jax.py\#L4-L6} and\newline\url{https://github.com/RobertTLange/evosax/blob/67b361/evosax/strategies/cma_es.py\#L97-L99}}
This demonstrates the readability of our code and its value for educational purposes.
Our library has introduced animated visualizations (Section~\ref{sec:vis}) that facilitate a more intuitive understanding of the behavior.
In addition to the simplicity of the code itself, our APIs are also simple and easy-to-use, which will be presented in Section~\ref{sec:getting_started}.
To continuously maintain the software quality, we have implemented fuzz testing to identify unexpected errors (Section~\ref{sec:fuzzing}).

To enhance \textbfit{practicality}, our library implements four methods that are highly relevant in practical scenarios.
The first method is for solving difficult problems such as multimodal and noisy issues \emph{without} the need for expensive hyperparameter tuning.
The second method is a transfer learning that accelerates CMA-ES by utilizing previous optimization results.
The third method is for handling mixed-variable domains, including integer and categorical variables rather than only continuous variables.
The fourth method is a multi-objective optimizer that supports mixed-variable domains.
The details are explained in Section~\ref{sec:recent}.
To the best of our knowledge, these methods have not yet been implemented in other evolution strategy libraries.
Therefore, their presence in our library represents a distinctive feature.
Importantly, these methods are generalizations of CMA-ES outlined in Section~\ref{sec:bg}, and thus do not require changes to the original CMA-ES implementation.
Consequently, our library retains its simplicity.
The API has also been designed for maximum simplicity, as will be detailed in Section~\ref{sec:recent}.

\subsection{Getting Started}
\label{sec:getting_started}

\paragraph{\textbf{Installation}.}
\texttt{cmaes} is available on the Python Packaging Index (PyPI)~\cite{pypi} and can be easily installed using the following command. Our library, written in pure Python and depending solely on NumPy~\cite{harris2020array}, is straightforward to install in a variety of user environments.
\begin{lstlisting}[]
$ pip install cmaes
\end{lstlisting}
For Anaconda users, the following command is recommended:
\begin{lstlisting}[]
$ conda install -c conda-forge cmaes
\end{lstlisting}

\paragraph{\textbf{Basic Usage}.}
To illustrate its basic functionality, Listing \ref{list:ask-and-tell-api} shows a simple example using \texttt{cmaes}.
The library employs an ask-and-tell interface~\cite{collette2013object}.
After initializing \code{CMA($\cdot$)} with initial \code{mean} and \code{sigma}, the process involves the following steps:
(1) Candidate solutions are generated using the \code{ask()} method.
(2) The sampled solutions are then evaluated by an objective function.
(3) The distribution parameters are updated based on the solutions and their objective function values using the \code{tell($\cdot$)} method.

\begin{lstlisting}[language=Python, caption=Simple example code with \texttt{cmaes}., label=list:ask-and-tell-api, numbers=left]
import numpy as np
from cmaes import CMA

def objective(x):
    # The best parameters are x = [3, -2]
    return (x[0] - 3) ** 2 + (10 * (x[1] + 2)) ** 2

optimizer = CMA(mean=np.zeros(2), sigma=2)
for generation in range(100):
    solutions = []
    for _ in range(optimizer.population_size):
        x = optimizer.ask()
        value = objective(x)
        solutions.append((x, value))

        print(f"{generation=} {value=} {x=}")
    optimizer.tell(solutions)
\end{lstlisting}

This ask-and-tell interface, by decoupling the optimizer from the objective function, adds flexibility to the library.
For example, it is easy to implement CMA-ES with restart strategies, such as IPOP-CMA-ES~\cite{auger2005ipop} or BIPOP-CMA-ES~\cite{hansen2009bipop}.
Listing~\ref{list:ipop-cma-es} provides an example code for IPOP-CMA-ES.
In this example, the algorithm restarts with a new mean vector and a doubled population size when the stopping conditions are met (i.e., when \code{should\_stop()} returns True).
These stopping conditions are based on those used in pycma~\cite{pycma}.
Additionally, we can specify box constraints by adding the argument, for example, \code{bounds=np.array([[-3,3],[-3,3]])} to \code{CMA($\cdot$)};
if a solution is not feasible, a resampling method is used, which is reasonable when the optimum solution is not very close to the constraints.

\begin{lstlisting}[language=Python, caption=Example code for IPOP-CMA-ES., label=list:ipop-cma-es, numbers=left]
import numpy as np
from cmaes import CMA

popsize = 6
optimizer = CMA(mean=np.zeros(2), sigma=3, population_size=popsize)

for generation in range(500):
    solutions = []
    for _ in range(optimizer.population_size):
        x = optimizer.ask()
        value = objective(x) # objective is omitted here
        solutions.append((x, value))
    optimizer.tell(solutions)

    if optimizer.should_stop():
        # Increase the popsize with each restart
        popsize = popsize * 2
        # Randomize the initial mean
        optimizer = CMA(
            mean=np.random.uniform(-5, 5, 2),
            population_size=popsize, sigma=3
        )
\end{lstlisting}

\subsection{Unexpected Error Identification via Fuzzing}
\label{sec:fuzzing}

To enhance software reliability, it is essential to proactively address unexpected error terminations.
These errors are often caused by unforeseen user inputs.
However, anticipating and listing all possible unexpected scenarios manually is a challenging and labor-intensive task.
This highlights the importance of automated methods, such as fuzzing, in simplifying this process.

\emph{Fuzzing} is an automated software testing method that aims to discover unexpected inputs that could cause program crashes.
It involves generating a wide range of inputs.
The key component in this process is the fuzzing engine, which automatically generates various inputs for the target program.
A notable technique is \emph{coverage-guided} fuzzing.
This approach not only generates random inputs but also collects code coverage data.
When an input is found to increase the code coverage, indicating untested paths, this information is used to methodically explore different branches within the program's code.
This strategy enhances the detection of unexpected errors, significantly contributing to the reliability of the software.

Listing~\ref{list:fuzzing} shows an example of fuzzing code used in our library.
In this example, to identify inputs that could lead to unexpected errors, fuzzing generates various inputs. These include evaluated solutions (\code{tell\_solutions}) for updating the distribution, the number of dimensions (\code{dim}), and the distribution parameters (\code{mean} and \code{sigma}).
To perform the coverage-guided fuzzing more efficiently, especially with the generation of NumPy arrays, we utilized Atheris~\cite{atheris} in combination with Hypothesis~\cite{maciver2019hypothesis}.

\begin{lstlisting}[language=Python, caption=Example code of fuzzing for finding the inputs that cause unexpected errors., label=list:fuzzing, numbers=left]
import atheris
from hypothesis import given, strategies as st
import hypothesis.extra.numpy as npst

@given(data=st.data())
def fuzzing_cmaes(data):
    # Generating random parameters.
    dim = data.draw(st.integers(min_value=2, ...))
    n_iterations = data.draw(st.integers(min_value=1))
    mean = data.draw(npst.arrays(shape=dim, dtype=float))
    sigma = data.draw(st.floats(min_value=1e-16))

    optimizer = CMA(mean, sigma)    
    for _ in range(n_iterations):
        # Checking "ask" function works properly
        optimizer.ask()
        
        tell_solutions = data.draw(st.lists(...))
        try:
            optimizer.tell(tell_solutions)
        except AssertionError:
            # Skip AssertionError exceptions, that are
            # raised if our generated data is invalid
            return

atheris.Setup(...)
atheris.Fuzz() # Brute forcing until the target crashes
\end{lstlisting}

\subsection{Animated Visualization}
\label{sec:vis}
We recognize the value of intuitively understanding the behavior of CMA-ES before delving into its detailed mathematical equations.
To achieve this, our library provides animated visualizations that illustrate the changes in the CMA-ES's multivariate Gaussian distribution.

\begin{figure}[htbp]
\centering
\includegraphics[width=0.43\textwidth,trim=5 25 5 20,clip]{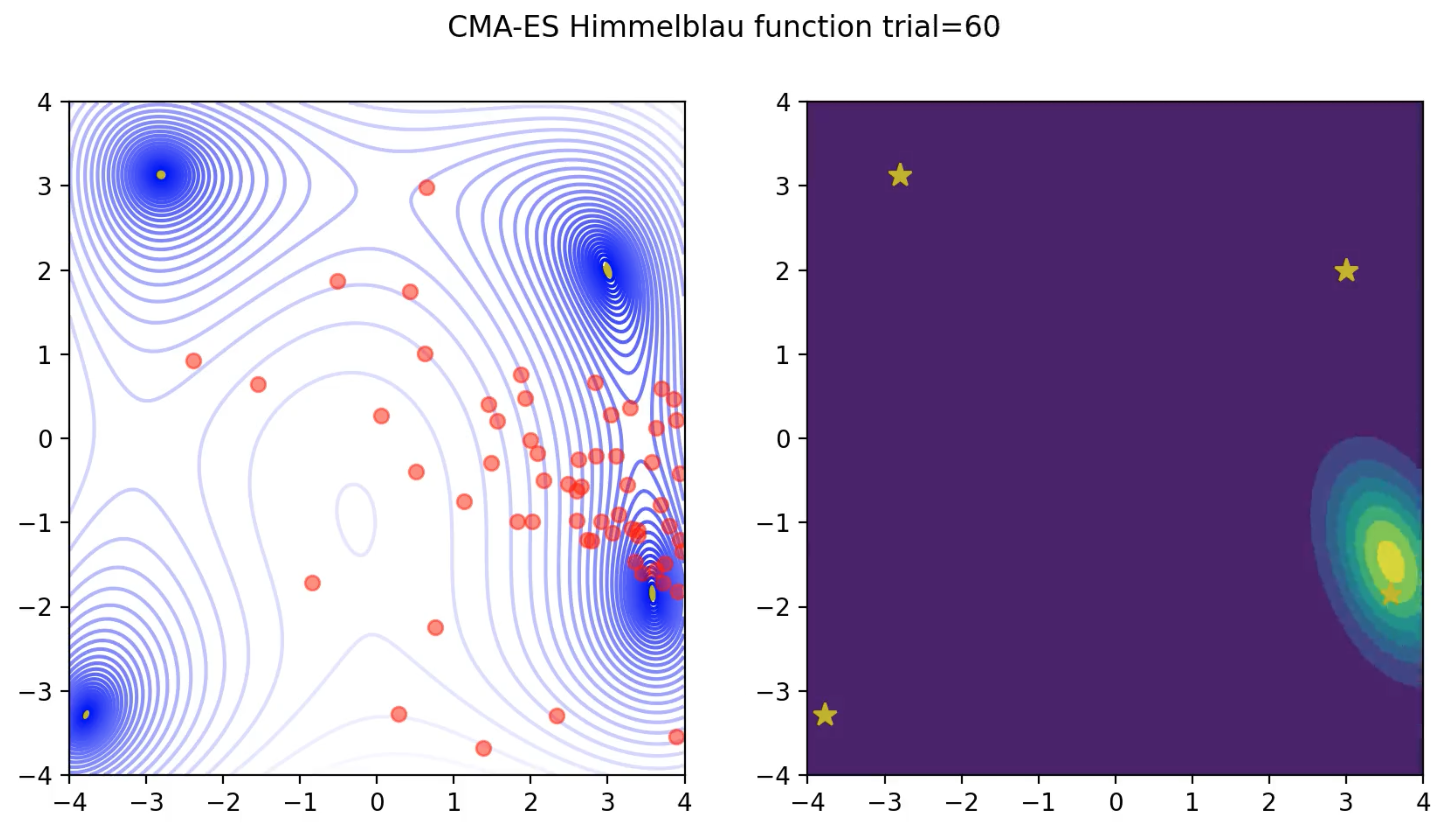}
\caption{Output example of animated visualization. (Left) Contour lines and sampled solutions. (Right) Multivariate Gaussian distribution in CMA-ES.}
\label{fig:cmaes-visualizer}
\end{figure}

Figure~\ref{fig:cmaes-visualizer} shows an example of the animated visualization for a single iteration.
Note that the actual output is an animated GIF, not a static figure, allowing for a dynamic observation of how CMA-ES progresses and converges towards the optimum.
This animated visualization aids in understanding the effect of various operations in CMA-ES, such as the rank-one update and step-size adaptation, in a visually interpretable way.
It is also beneficial for intuitively comparing behaviors across different optimization methods implemented in our library, such as separable CMA-ES~\cite{ros2008simple} and natural evolution strategies~\cite{wierstra2014natural}.
Consequently, animated visualization also serves as a useful tool for verifying the expected behavior of the implemented algorithms.

\subsection{Integration with Real-World Systems:\\ Use Case from Optuna}
\label{sec:optuna}
To exemplify the seamless integration capabilities of our library with other libraries, we present a use case involving Optuna~\cite{akiba2019optuna}, a renowned library extensively utilized for hyperparameter optimization of machine learning algorithms.
An example code of CMA-ES via Optuna is shown in Listing~\ref{list:optuna-cma-es-sampler}, where \code{CmaEsSampler} internally utilizes our library.

\begin{lstlisting}[language=Python, caption=Example code of Optuna with CmaEsSampler., label=list:optuna-cma-es-sampler, numbers=left]
import optuna

def objective(trial: optuna.Trial) -> float:
    x1 = trial.suggest_float("x1", -4, 4)
    x2 = trial.suggest_float("x2", -4, 4)
    return (x1 - 3) ** 2 + (10 * (x2 + 2)) ** 2

if __name__ == "__main__":
    sampler = optuna.samplers.CmaEsSampler()
    study = optuna.create_study(sampler=sampler, storage="sqlite:///optuna.db")
    study.optimize(objective, n_trials=250)
\end{lstlisting}

\begin{figure}[h]
\centering
\includegraphics[width=0.4\textwidth,trim=5 5 5 5,clip]{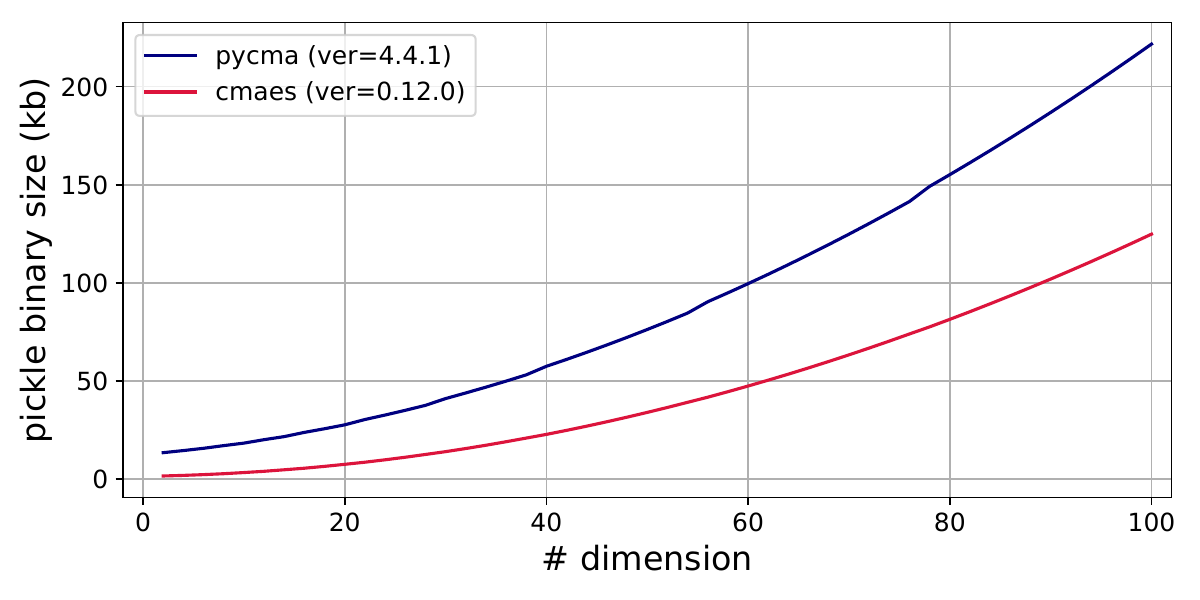}
\caption{Comparison of the serialization size for the CMA-ES objects after one generation using pickle in pycma and \texttt{cmaes}.}
\label{fig:pickle_size}
\end{figure}

A key factor in Optuna's integration of \texttt{cmaes} is not only its clean code and simple design but also the reduced size of its pickle serialization.
In hyperparameter optimization, the focus of Optuna, unexpected errors may occur during the training of machine learning models.
Consequently, it is crucial to make the optimization process resumable, necessitating the regular saving of CMA-ES optimization data, including distribution parameters.
Therefore, reducing the pickle size is critical for Optuna.
Our library has introduced specialized \code{\_\_getstate\_\_} and \code{\_\_setstate\_\_} methods that are invoked during pickle serialization and deserialization.
These modifications have successfully achieved further lightweighting of the pickle objects.
Figure \ref{fig:pickle_size} shows the reduction of the serialization size for the CMA-ES objects using pickle in \texttt{cmaes}, compared to pycma.

\section{Examples with Recent Advances}
\label{sec:recent}

A key aspect of our library is offering highly practical methods recently proposed in the context of CMA-ES through easy-to-use APIs.
In this section, we will highlight the following methods as examples, providing brief explanations and usage for each.
\begin{itemize}
    \item \textbfit{LRA-CMA}~\cite{nomura2023cma}: LRA-CMA can solve multimodal and noisy problems \emph{without} expensive hyperparameter tuning.
    \item \textbfit{WS-CMA}~\cite{nomura2021warm}: WS-CMA can solve a current task efficiently by utilizing results of a similar task and setting good initial distributions (i.e. transfer learning).
    \item \textbfit{CatCMAwM}~\cite{hamano2025catcmawm}: CatCMAwM can stably solve mixed-variable problems, including continuous, integer, and categorical variables.
    \item \textbfit{COMO-CatCMAwM}~\cite{hamano2025catcmatelo}: COMO-CatCMAwM uses multiple CatCMAwM instances to solve multi-objective mixed-variable problems.
\end{itemize}

Importantly, these methods are a generalization of CMA-ES and avoid unnecessary modifications to the original CMA-ES implementation.
This enables the preservation of the library's simplicity, making it easier for those who initially learned CMA-ES to straightforwardly delve into understanding these methods.
Beyond these, advanced methods such as Safe CMA-ES~\cite{uchida2024safe} and CMA-ES on sets of points (CMA-ES-SoP)~\cite{uchida2024sop} have been implemented in \texttt{cmaes}.

\subsection{Multimodal and Noisy Problems}

\begin{figure}[t]
    \centering
    \hspace*{\fill}
    \subfloat[][Rastrigin function \label{subfig:rastrigin}]{
        \includegraphics[width=0.49\linewidth]{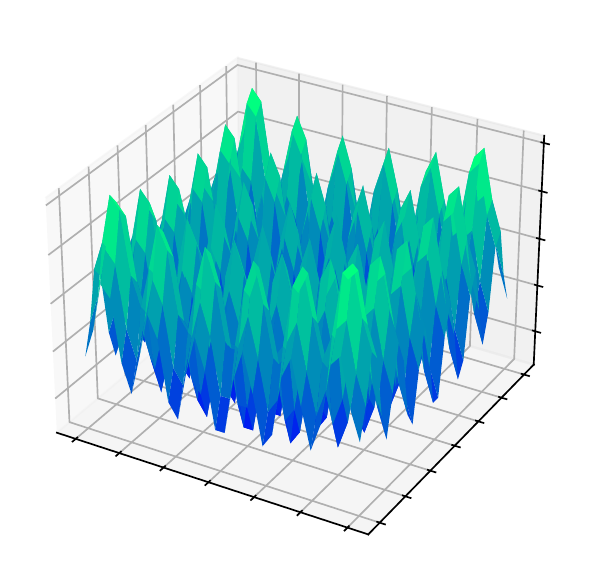}
    }
    \hspace*{\fill}
    \subfloat[][Noisy sphere function \label{subfig:noisy_sphere}]{
        \includegraphics[width=0.49\linewidth]{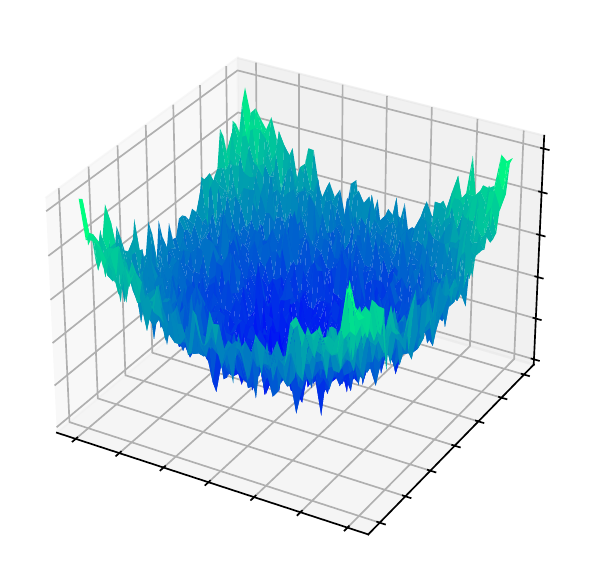}
    }
\hspace*{\fill}
\caption{Visualizations of function landscape. In noisy problems, although function value of each point is technically random variable, it can be treated as scalar value from optimizer's perspective in continuous space.
}
\label{fig:landscape}
\end{figure}

\paragraph{\textbf{Why do we need LRA-CMA\Qmark}}
The \emph{multimodality} of the objective function is one of the most important properties that make optimization difficult.
Figure~\ref{fig:landscape}\subref{subfig:rastrigin} shows the landscape of the Rastrigin function as an example of multimodal problems.
Due to the presence of numerous local optima in this function, CMA-ES may sometimes converge towards an incorrect solution.
A common practice to address these issues is to increase population size $\lambda$~\cite{hansen2004evaluating};
however, the hyperparameter tuning to determine the appropriate value for $\lambda$ often requires a significant amount of computational time and resources.

Another difficulty arising in practical applications is the presence of \emph{noise}, where the observed objective function value is obtained as $y = f(x) + \epsilon, \epsilon \sim \mathcal{N}(0, \sigma_n^2)$, with $\sigma_n^2$ representing the noise variance.
In this case, unless an appropriate $\lambda$ is chosen in accordance with the scale of the noise, attaining a high-quality objective function value becomes challenging.

Recognizing that increasing $\lambda$ has an effect similar to decreasing $\eta$~\cite{miyazawa2017effect}, our library employs automatic adaptation of the learning rate of the distribution parameters, i.e., $\eta$, rather than adaptation of the population size~$\lambda$~\cite{nishida2018psa} from the practical standpoint:
Practitioners often prefer to set $\lambda$ to a specific number of workers to prevent resource waste.
However, $\lambda$ adaptation might not always use all computing resources due to the variability of $\lambda$ during optimization.
In contrast, $\eta$ adaptation allows for maximal resource utilization by fixing $\lambda$ at the highest number of available workers.

\begin{figure}[tb]
\includegraphics[width=0.87\hsize,trim=5 5 5 5,clip]{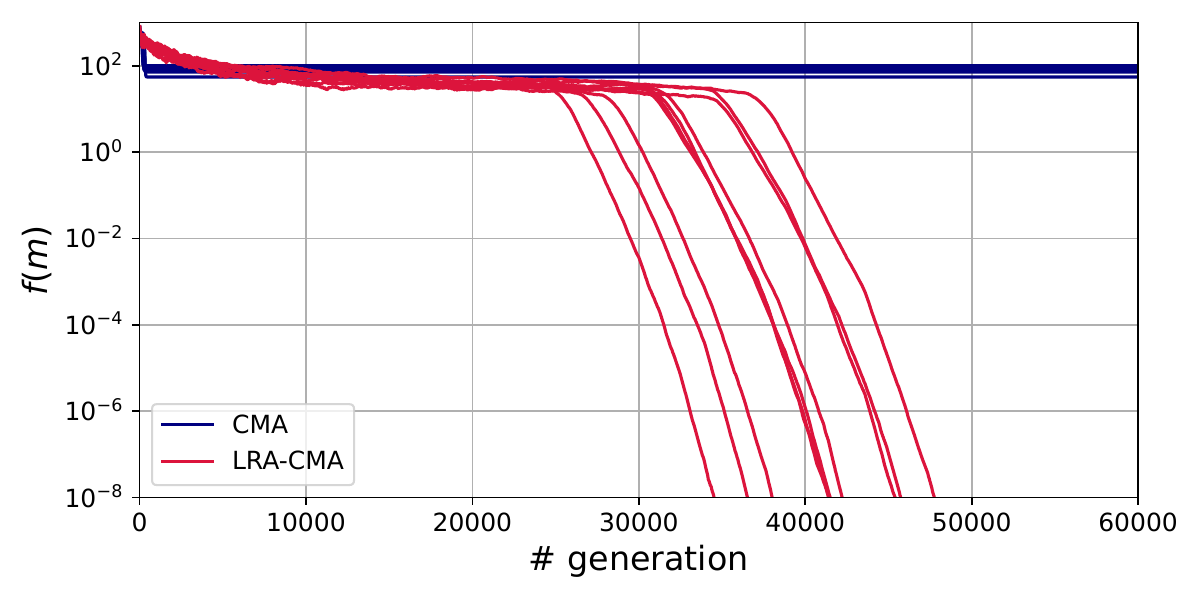}
\caption{Comparison between vanilla CMA-ES and LRA-CMA on the 40-D Rastrigin function.
Initial distribution is $m^{(0} = [3.0, \ldots, 3.0], \sigma^{(0)} = 2.0, C^{(0)} = I$.
Population size is set as the recommended value, i.e., $\lambda = 4 + \lfloor 3 \log (d) \rfloor = 15$. Vanilla CMA-ES gets trapped in local optima, whereas LRA-CMA succeeds in finding global optima \emph{without} any hyperparameter tuning.}
\label{fig:40d_rastrigin_comparison}
\end{figure}

\paragraph{\textbf{What is LRA-CMA\Qmark}}
Learning Rate Adaptation (LRA-)CMA~\cite{nomura2023cma,nomura2024cma} automatically adjusts $\eta$ to maintain a constant signal-to-noise ratio (SNR).
Therefore, $\eta$ decreases when SNR is low and increases when SNR is high, allowing for a responsive adaptation of $\eta$ to the search difficulty.
The validity of using SNR is discussed theoretically in \cite{nomura2024cma}.
Figure~\ref{fig:40d_rastrigin_comparison} shows the results of multiple trials on the Rastrigin function.
LRA-CMA, which includes $\eta$ adaptation, consistently finds the global optimum without any hyperparameter tuning, unlike the vanilla CMA-ES, which often gets trapped in local optima.

Notably, LRA-CMA is also effective in noisy environments.
This is based on the observation that noisy problems can be viewed as multimodal problems from an optimizer's perspective, as illustrated in Figure~\ref{fig:landscape}.
Indeed, LRA-CMA-ES has been observed to simultaneously address noise and multimodality issues effectively~\cite{nomura2023cma}.

\paragraph{\textbf{How to use?}}
Users can run LRA-CMA simply by adding the argument \colorbox[gray]{0.95}{\texttt{lr\_adapt=True}} during the initialization of \code{CMA($\cdot$)}.

\subsection{Transfer Learning}

\paragraph{\textbf{Why do we need WS-CMA\Qmark}}
Practically it is possible that we have already solved a problem (\emph{source task}) related to the problem we are currently trying to solve (\emph{target task}).
In such cases, it would be wasteful to optimize the target task from scratch without utilizing the results of the source task.
Therefore, it is expected that optimizing the target task can be made more efficient by transferring knowledge from the source task.
This knowledge transfer is especially beneficial when evaluating the objective function is expensive, such as in hyperparameter optimization~\cite{feurer2019hyperparameter}.

\begin{figure}[t]
    \centering
    \hspace*{\fill}
    \subfloat[][Collect solutions on\\ source task \label{subfig:a_source_solutions}]{
        \includegraphics[width=0.31\linewidth]{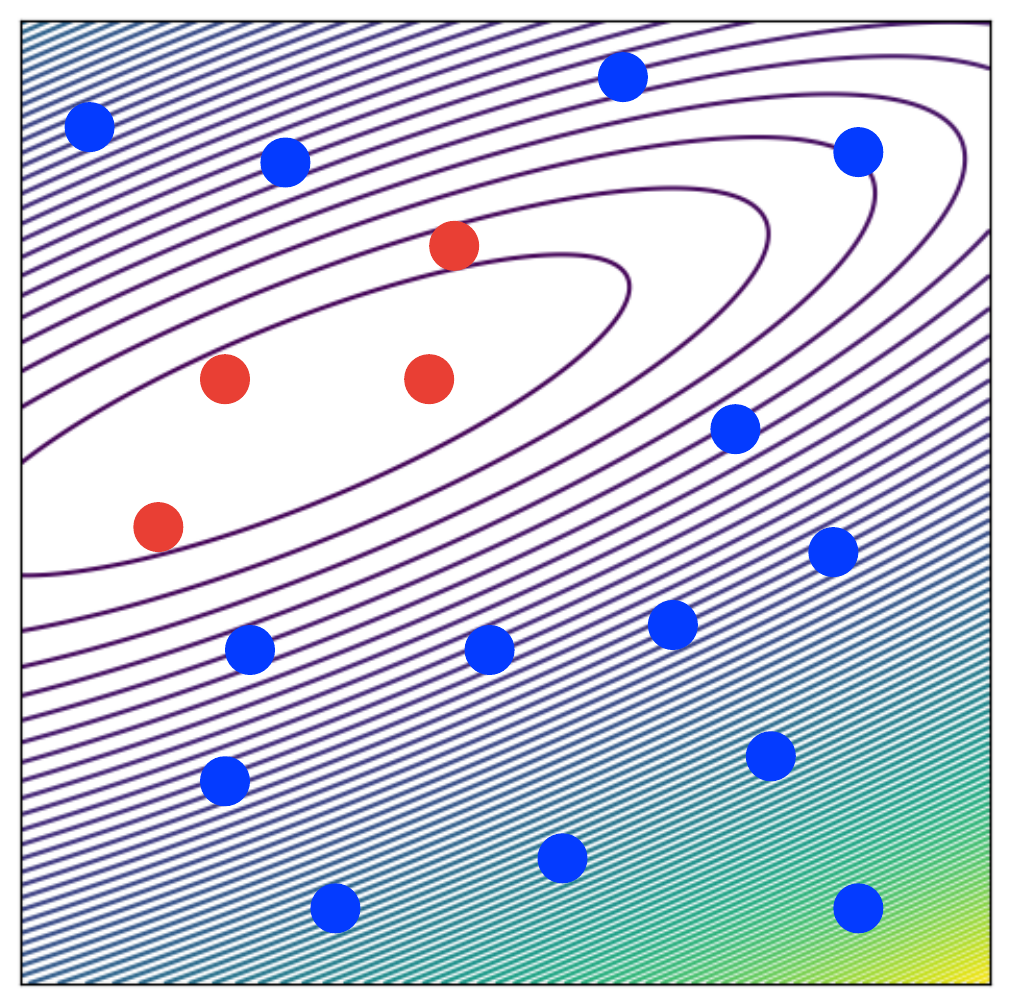}
    }
    \hspace*{\fill}
    \subfloat[][Estimate promising\\ distribution as GMM \label{subfig:b_promising_distribution}]{
        \includegraphics[width=0.31\linewidth]{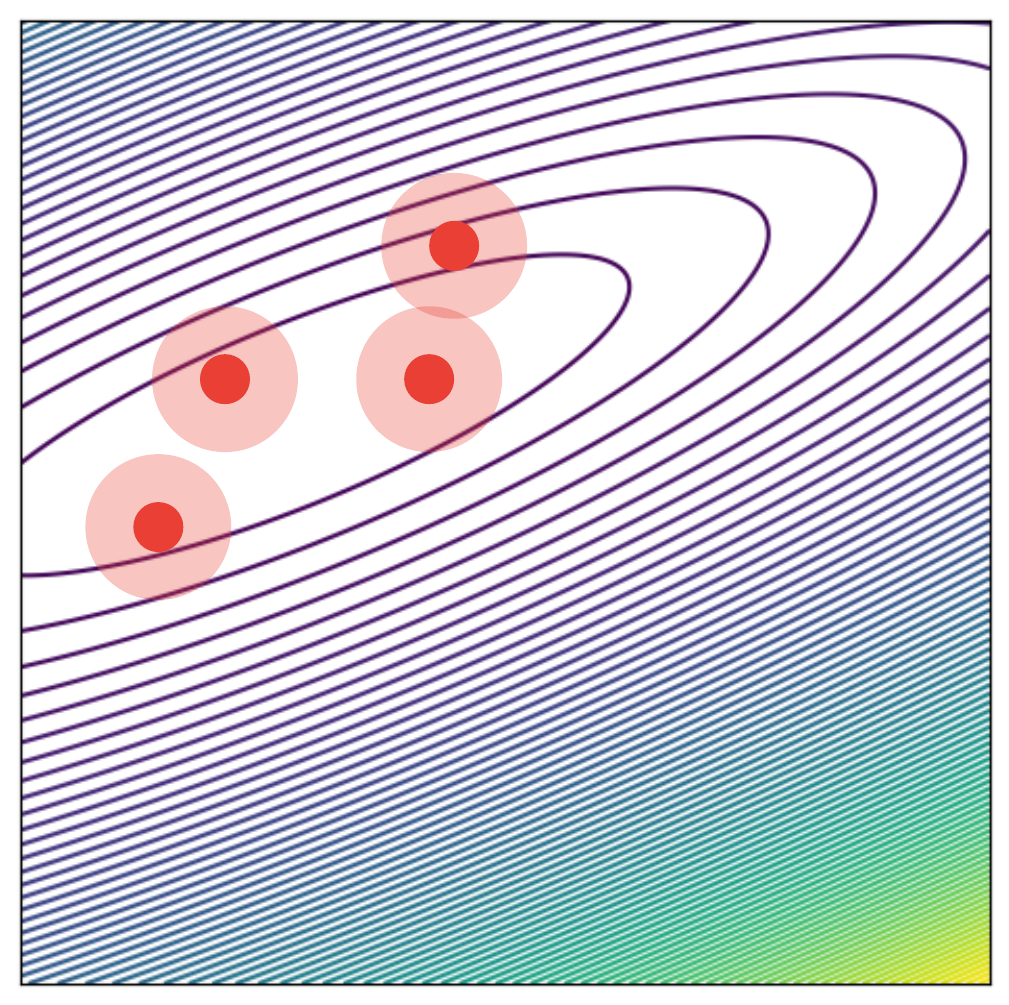}
    }
    \hspace*{\fill}
    \subfloat[][Initialize distribution\\ of CMA-ES for target task \label{subfig:c_initial_distribution}]{
        \includegraphics[width=0.31\linewidth]{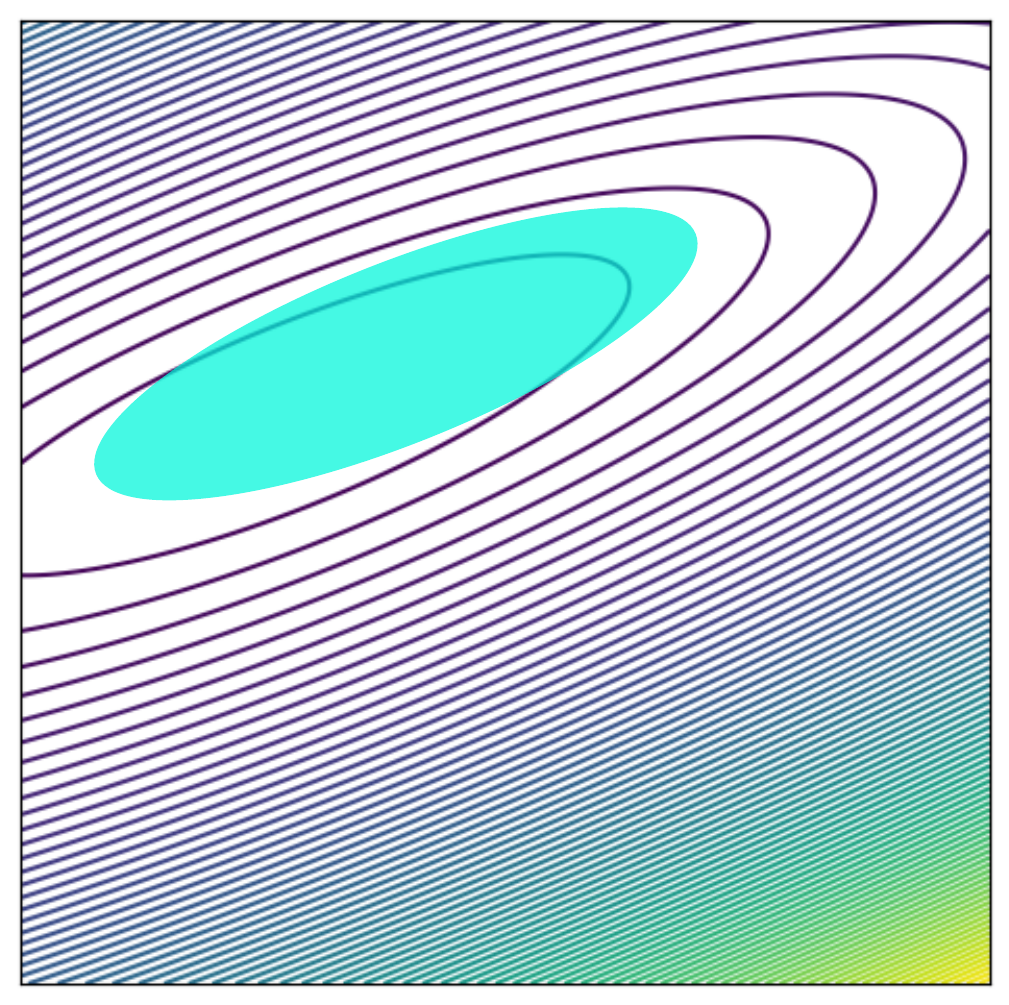}
    }\hspace*{\fill}
\caption{Procedure of WS-CMA.}
\label{fig:wscma_workflow}
\end{figure}

\paragraph{\textbf{What is WS-CMA\Qmark}}
Warm Starting CMA-ES (WS-CMA)~\cite{nomura2021warm} accelerates the optimization of the target task by utilizing information from the source task.
The process of WS-CMA is actually quite simple.
First, we assume that there are solutions evaluated on the source task (Figure~\ref{fig:wscma_workflow}\subref{subfig:a_source_solutions}).
Based on the results of the source task, WS-CMA first estimates a promising distribution as Gaussian mixture models (GMM) (Figure~\ref{fig:wscma_workflow}\subref{subfig:b_promising_distribution}).
It then determines the parameters of the initial distribution to fit the promising distribution by minimizing the Kullback-Leibler divergence between GMM and the initial distribution (Figure~\ref{fig:wscma_workflow}\subref{subfig:c_initial_distribution}).
Subsequently, WS-CMA executes CMA-ES using the determined initial distribution to efficiently solve the target task.

\paragraph{\textbf{How to use\Qmark}}
Listing~\ref{list:ws-cma} shows an example code using WS-CMA.
We first need to prepare an array (\code{source\_solutions} in the code) whose elements are tuples $(x, f(x))$.
By using the method \code{get\_warm\_start\_mgd($\cdot$)} with the created \code{source\_solutions}, we can obtain the \emph{promising} distribution parameters of the mean vector, the step-size, and the covariance matrix.
WS-CMA is then performed by initializing \code{CMA($\cdot$)} with these parameters.

\begin{lstlisting}[language=Python, caption=Example code of WS-CMA., label=list:ws-cma, numbers=left]
import numpy as np
from cmaes import CMA, get_warm_start_mgd

def source_task(x1: float, x2: float) -> float:
    return x1 ** 2 + x2 ** 2

def main() -> None:
    # Generate solutions from a source task
    # Please replace with your specific task when using
    source_solutions = []
    for _ in range(100):
        x = np.random.random(2)
        value = source_task(x[0], x[1])
        source_solutions.append((x, value))

    # Estimate a promising distribution
    ws_mean, ws_sigma, ws_cov = get_warm_start_mgd(
        source_solutions)
    optimizer = CMA(mean=ws_mean, sigma=ws_sigma, cov=ws_cov)
    ...
\end{lstlisting}

\subsection{Mixed-Variable Optimization}
\paragraph{\textbf{Why do we need CatCMAwM\Qmark}}
Many real-world black-box optimization problems involve mixed decision variables defined over a search space composed of the continuous domain $\mathcal{X}$, the integer domain $\mathcal{Z}$, and the categorical domain $\mathcal{C}$.
For example, hyperparameter optimization of deep learning models often requires jointly optimizing continuous variables (e.g., learning rates and regularization coefficients), integer variables (e.g., the number of layers), and categorical variables (e.g., choices of activation functions and optimization algorithms)~\cite{feurer2019hyperparameter,elsken2019neural}.
Similar mixed-variable optimization problems are also common in domains such as materials design and chemical reaction optimization~\cite{zhang2020bayesian,ozaki2020automated,aldulaijan2024adaptive}.

A straightforward method for handling discrete variables in CMA-ES is to map continuous samples to discrete values through rounding. However, as the multivariate Gaussian distribution converges, the generated discrete variables often become fixed to a (suboptimal) single value, leading to optimization stagnation. Furthermore, since categorical variables represent choices without an inherent numerical ordering, such discretization approaches impose unintended ordinal relationships that misrepresent the actual structure of the search space.

\paragraph{\textbf{What is CatCMAwM\Qmark}}
To address these challenges, our library implements CatCMA with Margin (CatCMAwM), a stochastic optimization method designed for continuous, integer, and categorical variables.
CatCMAwM is developed by incorporating novel integer handling into CatCMA~\cite{hamano2024catcma}, a mixed continuous-categorical black-box optimization method employing a joint distribution of multivariate Gaussian and categorical distributions.
Although CatCMAwM targets mixed-variable optimization, its integer handling is more effective than the original one used in CMA-ES with Margin (CMA-ESwM)~\cite{hamano2022cma, hamano2024marginal} even in mixed-integer optimization.
This ensures a stable search by imposing both lower and upper bounds on the marginal probabilities , preventing premature fixation of discrete values while maintaining the convergence efficiency of continuous variables.
On mixed-variable benchmarks, CatCMAwM achieves superior convergence performance compared to state-of-the-art Bayesian optimization methods.

\begin{figure}[t]
\centering
\includegraphics[width=0.35\textwidth]{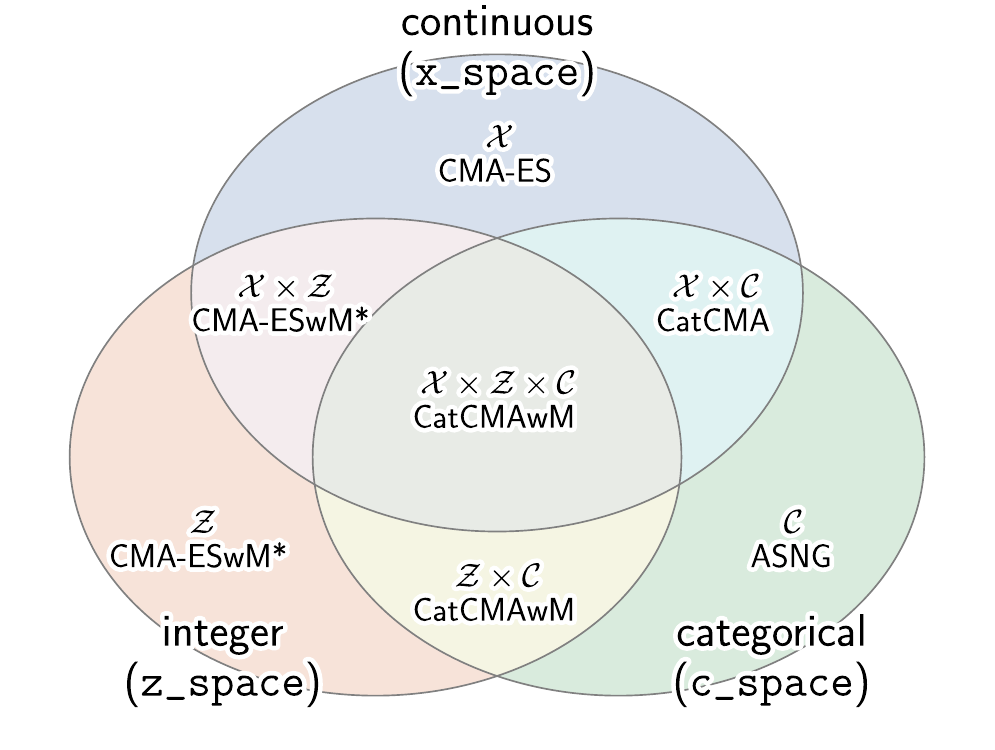}
\vspace{-2mm}
\caption{By partially setting the variable domains, continuous domain \code{x\_space}, integer domain \code{z\_space}, and categorical domain \code{c\_space}, users can run the appropriate algorithm within the \code{CatCMAwM} API \emph{without} tedious case-by-case switching. The asterisk (*) indicates that CatCMAwM uses revised integer handling derived from CMA-ESwM.}
\label{fig:venn-catcmawm}
\end{figure}

\paragraph{\textbf{How to use\Qmark}}
This CatCMAwM implementation provides a unified framework that automatically adapts to any combination of search spaces $\mathcal{X}$, $\mathcal{Z}$, and $\mathcal{C}$.
As demonstrated in Listing~\ref{list:catcmawm}, users can define the search space by specifying only the relevant arguments from \code{x\_space}, \code{z\_space}, and \code{c\_space}; unused domains can be simply omitted, allowing the framework to automatically configure the search distribution.
Figure~\ref{fig:venn-catcmawm} shows that the framework reduces to CMA-ES for continuous-only tasks, the Adaptive Stochastic Natural Gradient method (ASNG)~\cite{akimoto2019asng} for categorical-only tasks, CMA-ESwM for mixed continuous-integer tasks, and CatCMA for mixed continuous-categorical tasks.

As shown in the loop of Listing~\ref{list:catcmawm}, sampled solutions are returned as a \code{Solution} object, where specific variable sets are easily accessible via intuitive attributes such as \code{sol.x}, \code{sol.z}, and \code{sol.c}. 
This design allows users to focus solely on the relevant variable types for their problem while naturally ignoring unused domains.

\begin{lstlisting}[language=Python, caption={Example code of CatCMAwM on continuous, integer, and categorical optimization problems.}, label=list:catcmawm, numbers=left]
import numpy as np
from cmaes import CatCMAwM

def SphereIntCOM(x, z, c):
    ...

# [lower_bound, upper_bound] for each continuous variable
X = [[-5, 5], [-5, 5]]
# possible values for each integer variable
Z = [[-1, 0, 1], [-2, -1, 0, 1, 2]]
# number of categories for each categorical variable
C = [3, 3]

optimizer = CatCMAwM(x_space=X, z_space=Z, c_space=C)

for generation in range(50):
    solutions = []
    for _ in range(optimizer.population_size):
        sol = optimizer.ask()
        value = SphereIntCOM(sol.x, sol.z, sol.c)
        solutions.append((sol, value))
    optimizer.tell(solutions)
\end{lstlisting}

\subsection{Multi-Objective Optimization}
\paragraph{\textbf{Why do we need COMO-CatCMAwM\Qmark}}
Multi-objective optimization involves the simultaneous optimization of multiple, often conflicting, objective functions.
Its primary goal is to identify a diverse set of Pareto-optimal solutions.
For instance, multi-objective hyperparameter optimization~\cite{Hernandez2022mohpo, karl2023mohpo} targets objectives such as error-based performance, inference latency, and model size over the mixed-variable domain $\mathcal{X}\times\mathcal{Z}\times\mathcal{C}$.
While multi-objective extensions of CMA-ES have been proposed for continuous variables (MO-CMA-ES~\cite{igel2007mocma-1, igel2007mocma-2, voss2010improved-mocma}) and mixed-integer variables (MO-CMA-ES with Margin~\cite{hamano2024marginal}), these methods rely on elitist selection mechanisms, which are not straightforward to extend to categorical domains.

\begin{figure}[t]
\centering
\includegraphics[width=0.45\textwidth]{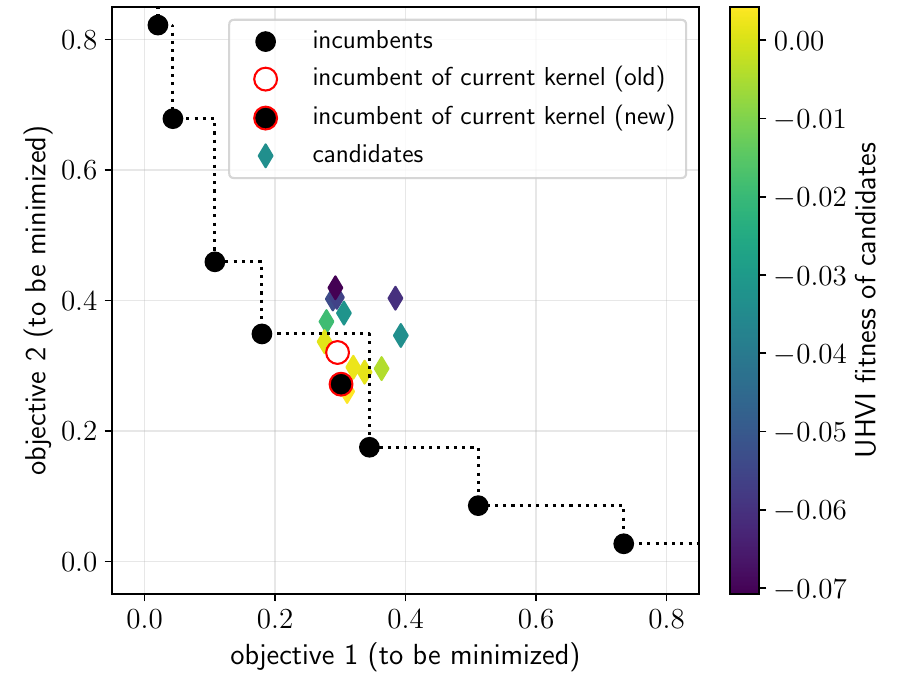}
\vspace{-1mm}
\caption{Illustration of the Sofomore framework update in the objective space. The framework optimizes kernels sequentially using UHVI fitness, which integrates the hypervolume contribution and the distance to the empirical Pareto front (dotted line) with respect to the remaining incumbents.}
\label{fig:sofomore}
\end{figure}

\paragraph{\textbf{What is COMO-CatCMAwM\Qmark}}
COMO-CatCMA with Margin (COMO-CatCMAwM) is a multi-objective extension of CatCMAwM based on the Sofomore framework~\cite{toure2019sofomore}.
This framework provides a general mechanism for constructing multi-objective optimizers by managing multiple single-objective algorithms, which are referred to as \emph{kernels}.
In COMO-CatCMAwM, each kernel is an instance of the CatCMAwM optimizer.
As illustrated in Figure~\ref{fig:sofomore}, these kernels are updated sequentially to optimize subproblems defined by the \emph{uncrowded hypervolume improvement} (UHVI) fitness, utilizing the update rule of CatCMAwM.
In bi-objective settings, COMO-CatCMAwM achieves competitive or superior hypervolume performance compared to established baselines such as NSGA-II~\cite{deb2002nsga2} and MOTPE~\cite{ozaki2020motpe, ozaki2022motpe}.

\paragraph{\textbf{How to use\Qmark}}
Listing~\ref{list:como-catcmawm} shows an example code using COMO-CatCMAwM.
Although COMO-CatCMAwM is technically applicable to problems with more than two objectives, our current implementation is limited to the bi-objective setting.
Similar to the CatCMAwM API, users can define mixed-variable search spaces intuitively.
Moreover, users do not need to track which internal kernel is currently being optimized and can simply use the \code{tell($\cdot$)} method, which automatically handles the sequential updates for the appropriate kernel.
\begin{lstlisting}[language=Python, caption=Example code of COMO-CatCMAwM on bi-objective mixed-variable optimization problems., label=list:como-catcmawm, numbers=left]
import numpy as np
from cmaes import COMOCatCMAwM

def DSIntLFTL(x, z, c):
    ...
    return [f1, f2]

# [lower_bound, upper_bound] for each continuous variable
X = [[-5, 15]] * 3
# possible values for each integer variable
Z = [range(-5, 16)] * 3
# number of categories for each categorical variable
C = [5] * 3

optimizer = COMOCatCMAwM(x_space=X, z_space=Z, c_space=C)

for _ in range(5000):
    sol = optimizer.ask()
    value = DSIntLFTL(sol.x, sol.z, sol.c)
    optimizer.tell((sol, value))
\end{lstlisting}

Unlike other methods that use a fixed population size, the Sofomore framework requires a variable number of evaluations per step.
This is because it evaluates both new candidates and current best solutions, called \emph{incumbents}.
To accommodate this variability, our library provides an iterator-based interface, \code{ask\_iter()}, which yields all candidates currently requiring evaluation:
\begin{lstlisting}[language=Python]
solutions = []
for sol in optimizer.ask_iter():
    value = DSIntLFTL(sol.x, sol.z, sol.c)
    solutions.append((sol, value))
optimizer.tell(solutions)
\end{lstlisting}
As shown above, the \code{tell($\cdot$)} method can also receive a list of solution and evaluation value pairs.
Notably, the order of submission to \code{tell($\cdot$)} is not required to match the sequence in which solutions were obtained from the ask interface.
This flexibility allows the optimizer to handle results as they become available, making it highly compatible with asynchronous parallel evaluations.

\section{Summary and Discussion}
\label{sec:conclusion}

This paper introduced \texttt{cmaes}, a simple yet practical CMA-ES library in Python.
The library's simplicity not only enhanced its software quality but also enabled easy integration with other libraries, such as Optuna~\cite{akiba2019optuna}.
Despite its simplicity, \texttt{cmaes} implements highly practical methods that were recently proposed, such as LRA-CMA~\cite{nomura2023cma}, WS-CMA~\cite{nomura2021warm}, CatCMAwM~\cite{hamano2025catcmawm}, and COMO-CatCMAwM~\cite{hamano2025catcmatelo}, all with easy-to-use APIs.
We hope that our library has become the ideal starting point for practitioners interested in exploring CMA-ES.

It is important to acknowledge that the features in our library are not comprehensive.
For example, pycma includes many features that our library lacks, as discussed in Section~\ref{sec:intro}.
Moreover, our library is not primarily designed for rigorous benchmarking tasks.
In this case, specialized software like COCO~\cite{hansen2021coco} is recommended.
In addition, the current library, including the implementation of COMO-CatCMAwM, is limited to the bi-objective case, even though the method is technically applicable to problems with more than two objectives.
Extending the implementation to general multi-objective settings is left for future work.

Finally, we openly and warmly invite discussions, questions, feature requests, and contributions to our library.
We believe that engaging in these interactions is crucial for identifying significant, yet unresolved, practical challenges.
It is our hope that our library will serve as a bridge between research and development, significantly enhancing the practical utility of CMA-ES.

\bibliographystyle{ACM-Reference-Format}
\balance
\bibliography{reference}
\end{document}